\documentclass[conference]{IEEEtran}
\IEEEoverridecommandlockouts
\usepackage{cite}
\usepackage{amsmath,amssymb,amsfonts}
\usepackage{graphicx}
\usepackage{textcomp}
\usepackage{xcolor}
\def\BibTeX{{\rm B\kern-.05em{\sc i\kern-.025em b}\kern-.08em
    T\kern-.1667em\lower.7ex\hbox{E}\kern-.125emX}}
\usepackage[utf8]{inputenc}
\usepackage[english]{babel}
 
\usepackage{amsthm}
\usepackage{algorithm}
\usepackage[noend]{algpseudocode}
\usepackage{float}
\usepackage{xcolor}
\usepackage{multirow}
\usepackage{subcaption}
\newcommand{\argmin}[1]{\underset{#1}{\operatorname{arg}\,\operatorname{min}}\;}

\newtheorem{thm}{Theorem}[section]

 \newtheorem{prob}[thm]{Problem}
\begin{document}
\title{Balancing Priorities in Patrolling with Rabbit Walks}
\author{Rugved Katole$^{1,2*}$, Deepak Mallya$^{1*}$, Leena Vachhani$^{1}$, Arpita Sinha$^{1}$
\thanks{$^{1}$ Systems and Control Engineering, Indian Institute of Technology Bombay, $^2$ Dept. of Mechanical Engineering, Birla Institute of Technology and Science Pilani, K.K. Birla Goa Campus}}
\maketitle
\begin{abstract}
    In an environment with certain locations of higher priority, it is required to patrol these locations as frequently as possible due to their importance. However, the Non-Priority locations are often neglected during the task. It is necessary to balance the patrols on both kinds of sites to avoid breaches in security. We present a distributed online algorithm that assigns the routes to agents that ensures a finite time visit to the Non-Priority locations along with Priority Patrolling. 
    The proposed algorithm generates offline patrol routes (Rabbit Walks) with three segments (Hops) to explore non-priority locations. 
    The generated number of offline walks depends exponentially on a parameter introduced in the proposed algorithm,
   thereby facilitating the scalable implementation based on the onboard resources available on each patrolling robot.
    A systematic performance evaluation through simulations and experimental results validates the proportionately balanced visits and suggests the proposed algorithm's versatile applicability in the implementation of deterministic and non-deterministic scenarios.
\end{abstract}


\section{Introduction}
Patrolling involves the systematic and repetitive visits of specific locations in a given environment. The task of patrolling plays a crucial role in the surveillance and monitoring of the environment, acting as a deterrent for anomalous activities. The multi-robot patrolling has been explored to consider repeated coverage with various patrol objectives. These methods have evolved to utilize multi-agent benefits, including cyclic strategies\cite{pasqualetti_optimal_2010}, partition-based strategies\cite{morales-ponce_optimal_2022, portugal_msp_2010}, learning-based patrolling \cite{caccavale_multi-robot_2023, chen_multi-agent_2021}, adversarial patrolling \cite{talmor_power_2017}, and negotiation mechanisms\cite{hwang_cooperative_2009}.

It has been shown that Static surveillance using CCTV reduces the incidence of theft and burglaries by 19\% \cite{malpani_impact_2020}. Furthermore, crime rates have reduced by 23\% during police patrolling \cite{ratcliffe_philadelphia_2011}. Additionally, robots equipped with sensors are used for regular inspection and monitoring of infrastructure like buildings, bridges, tunnels, storage tanks, pipelines, and roads. They enhance the safety of operations and improve cost-effectiveness \cite{halder_robots_2023}. The global market for inspection robots is projected to reach \$18.9 billion by 2030 \cite{ltd_inspection_2023}.

The patrolling environment contains different patrolling locations, with certain locations of higher priority than others. The locations of higher priority (Priority Nodes) are usually the critical points of threats or failures. Although these priority locations are of absolute importance, patrolling other locations (Non-Priority Nodes) is also necessary to avoid disruption in operation. From a security standpoint, if the Non-Priority Nodes are not patrolled frequently, it makes it easier for adversaries to infiltrate through these locations. In this work, we present a patrolling methodology that balances the visits to priority and Non-Priority Nodes. The challenge is in associating parameters that would ensure the desired balancing.


The notion of optimality is to minimize the overall graph idleness or worst idleness, i.e., the time delay between consecutive visits to a location \cite{almeida_recent_2004,portugal_survey_2011,huang_survey_2019,basilico_recent_2022}. 
Typically, the environment is represented using an undirected graph with nodes representing visiting locations and edges representing connecting paths. The edge weights are characterized by factors such as distance \cite{pippin_performance_2013, pasqualetti_optimal_2010}, expected travel time \cite{farinelli_distributed_2017}, priorities \cite{chen_fast_2012}, and visitation frequency \cite{sea_frequency-based_2018}. These weights play a crucial role in decision-making by assigning different rewards to visit different nodes. Usually, reward functions in multi-robot patrolling employ idleness, which measures the time elapsed between two consecutive visits, as a key component. The optimality of an algorithm is often assessed by minimizing the overall idleness. Different multi-agent architectures have been evaluated for patrolling under varying parameters, including reactive algorithms versus cognitive algorithms and coordination schemes, using different metrics based on idleness \cite{machado_multi-agent_2003}.

 A factor approximation to the optimal solution for cyclic, acyclic, and chain graphs through partitioning has been studied in the past \cite{pasqualetti_optimal_2010,afshani_approximation_2021}. Heavy-edge heuristics are used in this multi-level graph partitioning approach for consistent partitioning of large graphs \cite{portugal_msp_2010}. The method achieves this through multi-node swapping based on equilibrium, with the goal of reducing redundancy and minimizing robot detritions. The k-means clustering for partitioning and a Simulated Annealing algorithm \cite{sea_frequency-based_2018} have been developed for pathfinding. Lauri et al. \cite{lauri_two-step_2008} uses ant colony optimization techniques for partitioning and path-finding. These centralized methods have a central station/node for partitioning and calculating the shortest paths for a robot to patrol. However, the centralized approach is prone to a single point of failure. Thereby, researchers have explored distributed approached for a robust alternative. 

The distributed approach typically uses reward-based online route assignment, the rewards functions use idleness values along with other metrics to achieve optimality. The work in \cite{almeida_combining_2003} extends \cite{machado_multi-agent_2003} by deepening the heuristics and choosing a slightly longer path to reduce overall idleness. 

Other multi-agent patrolling methods include auction-based patrolling and learning-based patrolling. In the auction-based method, the robots negotiate a bid for a node, and through communication, the best robot gets to patrol the node. The bidding is based on different heuristics such as patrol path length \cite{hwang_cooperative_2009}, path distance \cite{pippin_performance_2013}, travel cost, and number of tasks \cite{farinelli_distributed_2017}. 

The learning-based methods are highly suited for dynamic environments due to their adaptability and probabilistic nature, the robots use their local idleness values to make local patrolling decisions. A patrolling problem modeled as a Semi-Markov decision process for cooperative multi-agent reinforcement learning. The agents communicate using flags or through broadcasting intentions or both, a rewards function that maximizes selfish utility is found to be more effective than the one using wonderful life utility (utility considering all other agent intentions) \cite{santana_multi-agent_2004}. A recent work leverages graph attention networks for persistent monitoring using multi-agent reinforcement learning \cite{chen_multi-agent_2021}. The agents share locally perceived information using graph attention networks, a neural network architecture to operate on graph data structure \cite{velickovic_graph_2018}. In \cite{portugal_cooperative_2016, portugal_applying_2013}, authors use Bayesian reasoning and learning to adapt to system dynamics. It uses a Bayesian decision-making model with likelihood reward-based learning and continued prior updates.

All the work in multi-agent patrolling leads to either deterministic or non-deterministic patrolling routes. In a scenario where adversarial forces attempt to infiltrate the patrolled area, the knowledge of patrolling routes can be easily obtained by observing the agents over a period of time. For problem objectives with patrolling locations of absolute importance (priority), it is also important to sweep the other environment locations as frequently as possible to eliminate threats from those areas. Previously, this has been addressed through weighted graphs \cite{alamdari_persistent_2014, afshani_approximation_2021}, where the routes are generated based on the ratio of maximum and minimum weights of locations (priority and Non-Priority Nodes, respectively, in the graph). 
A solution based on route selection has a clear advantage of faster sweeping of the environment as compared to one that is based on node selection. 

A reward-based route selection algorithm \cite{mallya_priority_2021} has been developed for priority patrolling. 
The reward function considers idleness overshoot from the given time period of visits for Priority Nodes, their proximity, and the length of the route for online route assignment.
A partitioning algorithm \cite{morales-ponce_optimal_2022} based on the distribution of priority and Non-Priority Nodes is also used for priority patrolling. Each segment is then patrolled by a robot. 
Deep Q-learning is employed for prioritizing sanitation of crowded spots at railway stations \cite{caccavale_multi-robot_2023}.

Also, the number of routes to explore increases exponentially with the time period, and an exhaustive search that considers all the routes for selection is impractical. 
The deterministic strategy is well suited for monitoring or inspection purposes where the threat of attack is not considered. On the other hand, security and surveillance purposes require non-deterministic strategies to ensure safety. Moreover, the patrolling problem focuses on balancing the visits to all the locations/nodes, while the priority patrolling problem aims to visit the Priority Nodes, and visits to Non-Priority Nodes are unaccounted for. A distributed strategy that ensures priority patrolling with accountability on visits to all locations (including non-priority ones).

In this paper, we focus on multi-agent distributed patrolling to balance visits between different priority and non-priority locations in an environment. 
The objective is to visit priority locations as well as non-priority locations such that the Non-Priority Nodes are visited in a finite time. The algorithm explores more routes with Non-Priority Nodes such that the trade-off between the increase in priority location idleness is compensated by the reduction in overall graph maximum idleness. The algorithm is designed for practical implementation on robots with limited resources. The algorithm uses a parameter based on the robot's memory for route generation and further selects a route from a subset of the generated routes. This reduces the computational and memory requirements of the robot. 

\textbf{Contributions}: The contributions of this work are summarized as follows.
\begin{itemize}
    \item A new approach for efficiently balancing priorities among priority and non-priority locations/nodes during patrolling tasks. The proposed algorithm dynamically assigns routes to robots, minimizing the delay between subsequent visits to each location. 
    \item A novel resource-aware route generation algorithm. The algorithm generates a finite number of routes based on a tunable parameter, enabling exploration within the robot's onboard resource constraints. 
    \item Performance trade-offs between the proposed four variant algorithms of online robots' route assignments designed to balance priorities are obtained. These deterministic and non-deterministic variants ensure that the visits to the Priority and Non-Priority nodes are proportionately balanced.
    
\end{itemize}

The remaining paper is organized as follows: In section \ref{Problem_formulation}, we formalize the Priority Patrolling Problem (PPP) for proportionately balanced visits and establish the terminology. Section \ref{PPA} provides the four variants of solution to PPP for resource-aware implementations. In section \ref{setup}, we present the evaluation metrics and discuss systematic simulation and experimental analysis of proposed variants. Finally, section \ref{conclusion} concludes the paper and presents future directions.

\begin{figure}
    \centering
    \fbox{\includegraphics[width = 0.48\textwidth]{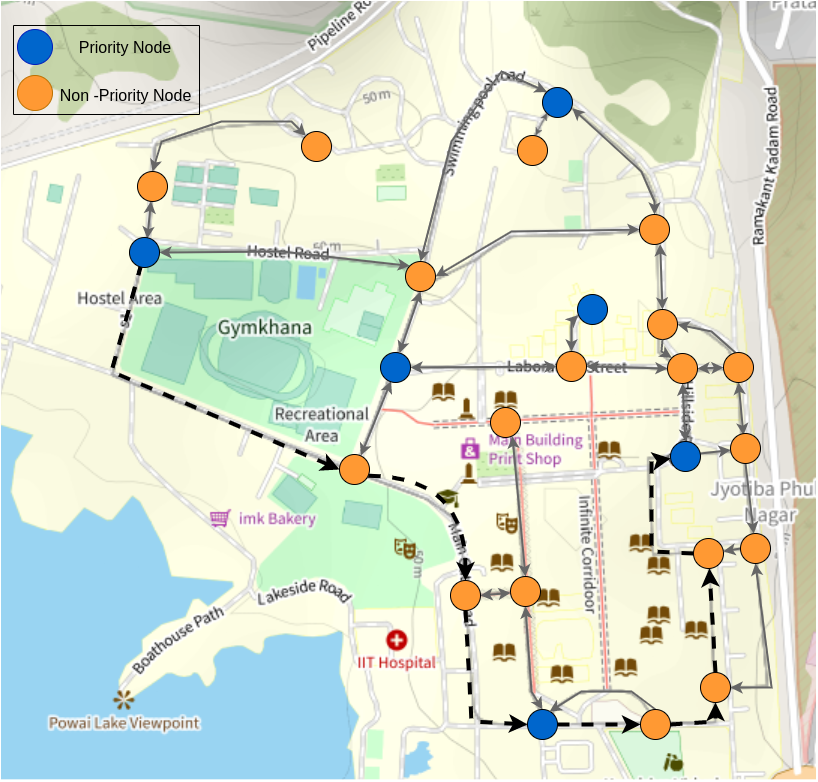}}
    \caption{A Graph $\mathcal{G}(\mathcal{V}, \mathcal{E})$ representing a patrolling environment with Priority (blue colored circles) and Non-Priority Nodes (orange colored circles). The dashed line represents a walk between two priority nodes}
    \label{fig: Nodes}
\end{figure}


\section{Problem Formulation}
\label{Problem_formulation}
The patrolling environment is represented as a strongly connected directed graph denoted using $\mathcal{G}(\mathcal{V}, \mathcal{E})$, where $\mathcal{V}$ is a set of nodes indicating patrolling locations and $\mathcal{E}$ is the set of edges describing the road segment between two nodes $v_i$ and $v_j$. 
Given a patrolling environment $\mathcal{G}(\mathcal{V}, \mathcal{E})$, the objective of the priority patrolling problem is to visit the Priority Nodes as often as possible and also ensure visits to Non-Priority Nodes. 
The set of \emph{Priority Nodes} is denoted by $\mathcal{S} \subset \mathcal{V}$ and the set of remaining \emph{Non-Priority Nodes} is denoted by $\mathcal{NS}  = \mathcal{V} \setminus \mathcal{S}$. The set of identical mobile agents is denoted by $\mathcal{A} = \{a_1, \dots, a_k\}$. Figure \ref{fig: Nodes} shows an illustrative graph with Priority and Non-Priority Nodes. The {Patrol Strategy} is denoted by $\mathcal{P} = \{w_1, \dots, w_k\}$ where $w_i \in \mathcal{P}$ is the walk (a sequence of non-repetitive consecutive edges) traversed by an agent $a_i \in \mathcal{A}$ during the patrol. 

At the time $t$, {Instantaneous Idleness} (or, simply {Idleness}) $I_i(t)$ of a node $v_i \in \mathcal{V}$ is the time elapsed since the last visit to $v_i$ by an agent $a_i \in \mathcal{A}$. The {Instantaneous Maximum Idleness} value for any node amongst the {Priority Nodes} is given by \[\mathcal{I}_S(t) = \max_{v_i \in \mathcal{S}} I_i(t)\]

Now, the objective is to reduce the Maximum Idleness of Priority Nodes while patrolling the entire environment such that the idleness of Non-Priority Nodes is finite.

\begin{prob}[\textbf{PPP}: Priority Patrolling Problem]
    Given a patrolling environment represented as {Graph} $\mathcal{G}(\mathcal{V}, \mathcal{E})$ with {Priority Nodes} $\mathcal{S}$ and the set $\mathcal{A}$ of agents, find a {Patrol Strategy} $\mathcal{P}$ that achieves \[\min_{\mathcal{P}} \max \mathcal{I}_S(t)\] such that maximum idleness of each {Non-Priority Node} is finite. 
    \label{prob:pp}
\end{prob}


\section{Priority Patrolling Algorithm}
\label{PPA}

The Priority Patrolling Algorithm aims to find an optimal strategy for each agent to minimize the maximum Priority Node idleness such that the idleness of all Non-Priority Nodes is finite. For each agent, the algorithm selects a walk from one Priority Node to another to achieve the balanced visits of all nodes. 
As an agent completes its walk, it is assigned another walk. There are two phases to finding an optimal strategy - offline and online. In the first phase (offline), we compute the walks from one Priority Node to another Priority Node. For an exhaustive search strategy, the number of walks generated increases exponentially with the size of the graph. We address this issue by introducing ``Rabbit Walks,'' each consisting of three ``Hops'' (analogous to a rabbit's hop). 

In the second phase (online), we select a target Priority Node and a Rabbit Walk that maximizes the sum of Idleness of nodes.

\subsection{Phase 1: Rabbit Walks Generation}
We define a Rabbit Walk as a walk between two Priority Nodes, $v_s$ and $v_t$, with three consecutive Hops.\footnote{The source Priority Node $v_s$ and target Priority Node $v_t$ may or may not be the same.}
The first Hop is a walk starting from source node $v_s$ and has ``H'' nodes. The next two Hops are the shortest paths from the end of the first Hop to the Priority Node $v_t$ via a random node $v_R$. 

\begin{figure}[htbp]
    \centering
    \includegraphics[width = 3.2in]{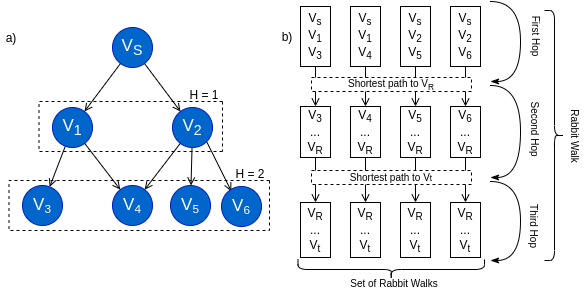}
    \caption{a) Illustration of a tree constructed from the graph $\mathcal{G}$. b) Each path from the tree is extended to $v_t$ via $v_R$ to form the three Hops of a Rabbit Walk}
    \label{fig: RWG} 
\end{figure}

The Rabbit Walk Generator Algorithm (\ref{alg: RWG}) is divided into three steps. In step 1 (Lines 1-7), we construct a tree with depth $H$ and the source Priority Node $v_s$ as the root of the tree. The children of each node in the tree are its neighbors $\mathcal{N}$ in Graph $\mathcal{G}$. To avoid cyclic walks (a walk with repetitive edges), we ensure the pair of the last node in the candidate walk $w.last$, and the new node $v$ is not an edge in $w$ (Line 5). In Line 6, we extend each walk $w$ with $h$ nodes (a walk in $W_h$) with a node $v$ and store it in the set of walks $W_{h+1}$. 
For instance, Figure~\ref{fig: RWG}(a) illustrates a tree with depth $H = 2$. The tree starts with the source Priority Node $v_s$, which then extends to its neighbors at $H = 1$ and to neighbors of neighbors at $H = 2$. 
Each path from the source to the leaf of the tree represents a candidate walk to be considered for the first Hop. The number of candidate walks depends on the depth $H$ and the $d$ degree of the graph. Hence, at most, $d^H$ walks will be generated in the first step. 

\begin{algorithm}[htbp]
    \caption{Rabbit Walk Generator}
    \textbf{Input:} $H$, $\mathcal{G}$, $v_s$, $\mathcal{S}$\\
    \textbf{Output:} $\{W_{s}^t\}$ \Comment{Set of walks from $v_s$ to $v_{t}$, for each $v_t \in S$}\\
    \textbf{Initialize:} $w \leftarrow (v_s), h \leftarrow 0, W_h \leftarrow \{w\}, W_s^t \leftarrow \emptyset$
    \begin{algorithmic}[1]
        \While {$h < H$}
        \State $W_{h+1} \leftarrow \emptyset $
            \For {$w \in W_h$}
                \For {$v \in \mathcal{N}(w.last) $} \Comment{Explore neighbors}
            \If{$pair(w.last, v) \notin w$}
                \State $W_{h+1} \leftarrow W_{h+1} \cup \{(w+v)\}$
            \EndIf
            \EndFor
            \EndFor
            \State $\emph{h \small{++}}$ 
        \EndWhile
        \For{$w \in W_{H}$}
            \For{$v_R \in \mathcal{V} \setminus \{w\}$}
                \State $w \leftarrow w +(shortest\_path(w.last,v_R))$
                \For{$v_{t} \in S$}
                    \State $W_s^t \leftarrow W_s^t \cup \{w +(shortest\_path(v_R,v_t))\}$
                \EndFor
            \EndFor
        \EndFor
    \Return $\{W_{s}^1,W_{s}^2,\dots,W_{s}^{\vert\mathcal{S}\vert}\}$
    \end{algorithmic}
    \label{alg: RWG}
 \end{algorithm}  

For the second and third Hop, a random node $v_R$ is selected from the nodes not in $w$, and a candidate walk from the first step. 

Step 2 (lines 8-10) of the algorithm extends $w$ by the shortest path between the last node $w.last$ to node $v_R$(line 10). 
Similarly, in step 3 (lines 11-12), the shortest path from $v_R$ to the target Priority Node $v_t$ (line 12) generates the third Hop. 
These walks are then stored in the set $W_s^t$ for $v_t \in \mathcal{S}$ and the algorithm returns a collection of all walks starting from source node $v_s$ to all Priority Nodes $v_t \in \mathcal{S}$.

\textit{Remark 1:} The maximum number of Rabbit Walks generated is the product of the number of candidates for First, Second, and Third Hops given by $d^H \times (|\mathcal V|- H+1) \times |\mathcal S|$.
\subsection{Phase 2: Rabbit Walk Assignment}
A Rabbit Walk is assigned to an agent at source node $v_s$ from the set of walks obtained in Phase 1. The sets of Rabbit Walks are arranged as $W_s^t$, walks between all pairs of Priority Nodes $v_s$ and $v_t$.
To avoid parsing through all the walks exhaustively, we propose four different ways to select a subset of walks $W_s$ from the previously generated rabbit walks between all priority nodes. 
\begin{itemize}
    \item PPA-Exhaustive: Every Priority Node is a candidate Target Node. $$W_{s} = \bigcup_{v_t \in \mathcal{S}} W_{s}^t$$
    \item PPA-Sampled: Based on an additional parameter $N$, we sample $N$ Priority Nodes from $\mathcal{S}$ and consider them as Target Nodes. $$ W_{s} = \bigcup_{v_t \in f(\mathcal{S}, N)} W_{s}^t$$ where, $f(\mathcal{S}, N)$ denotes the $N$ nodes sampled from $\mathcal{S}$ at the time of assignment.
    \item PPA-Random: In this case, we select one Priority Node from $\mathcal{S}$ arbitrarily and consider it as the Target Node. $$v_T \sim Uniform(\mathcal{S}); \;\;W_s = W_s^T$$
    \item PPA-Greedy: We maintain a counter $C_t$ for each Priority Node $v_t$, denoting the number of times the corresponding node is assigned as the Target Node. We then select the Priority Node with the least count as the Target Node. $$v_T = \argmin{v_t \in \mathcal{S}}{C_t};\;\; W_s= W_s^T$$
\end{itemize}
\textit{Remark 2:} The PPA-Exhaustive searches all the Rabbit Walks, i.e., at most $d^H \times (|\mathcal V|- H+1) \times |\mathcal{S}|$ walks. Whereas PPA-Random and PPA-Greedy search at most $d^H \times (|\mathcal V|- H+1)$. The PPA-Sampled searches $d^H \times (|\mathcal V|- H+1) \times N$ Rabbit Walks.

Next, the reward calculation for a walk $w \in W_s$ is given in Equation~\ref{eq1}. 
\begin{equation}
     Reward(t) = \sum_{v_j \in w} I_j(t),
\label{eq1}
\end{equation} 
where, $I_j(t)$ is the instantaneous idleness of node $v_j \in \mathcal{V}$.

The agent is assigned the walk $w$ with the maximum Reward value to solve Problem~\ref{prob:pp}. In summary, the proposed method of walk assignment obtained through $W_s$, a subset of Rabbit Walks ensures frequent visits to Priority Nodes while guaranteeing visits to Non-Priority Nodes in finite time.


\section{Simulations and Experimental results}
\label{setup}
In this section, we present the results of our extensive simulations focused on evaluating the performance of the patrolling algorithm across various scenarios and layouts. The proposed algorithm and its variants are evaluated using the following metrics at the end of each simulation:

\begin{itemize}
    \item Priority Nodes' Maximum Idleness: Indicates the maximum time a Priority Node was idle or unvisited.
    $$ \max_{v_i \in \mathcal{S}} I_i(t)$$
    \item Graph Maximum Idleness: Indicates the maximum time any node was idle or unvisited.
    $$ \max_{v_i \in \mathcal{V}} I_i(t)$$
    \item Idleness Ratio: The ratio of Graph Maximum Idleness to the Priority Nodes' Maximum Idleness. It indicates the proportional priority given to the Non-Priority Nodes and a measure for balanced visits.  
    $$\frac{\max_{v_i \in \mathcal{V}} I_i(t)}{\max_{v_i \in \mathcal{S}} I_i(t)} $$
\end{itemize}
\subsection{Simulation Settings}
The developed algorithm is evaluated through comprehensive simulations on Simulator for Urban MObility (SUMO) \cite{lopez_microscopic_2018}. SUMO is an open-source traffic simulator capable of large-scale traffic simulations. It hosts a number of vehicle types and motion models. The motion planning is handled by in-built functions, allowing effortless evaluation of patrolling strategies. The Traffic Control Interface (TraCI) is a Python API for SUMO for real-time control of vehicles. During the patrol, agents communicated with the server to track the idleness of the node. The maximum velocity of a patrolling vehicle is $10 m/s$, and each simulation runs for 20,000 seconds. 

\subsection{Graphs and other Settings}
\begin{figure}[htbp]
\centering
\fbox{\begin{subfigure}[b]{0.17\textwidth}
    \includegraphics[width=\textwidth]{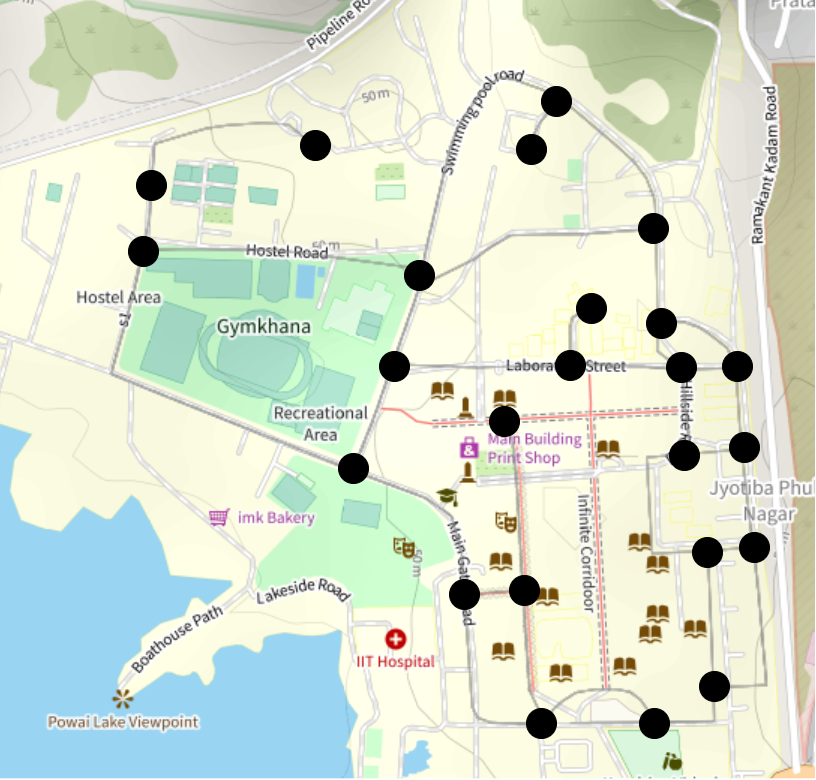}
    \caption{IIT Bombay}
\end{subfigure} 
\begin{subfigure}[b]{0.115\textwidth}
    \includegraphics[width=\textwidth]{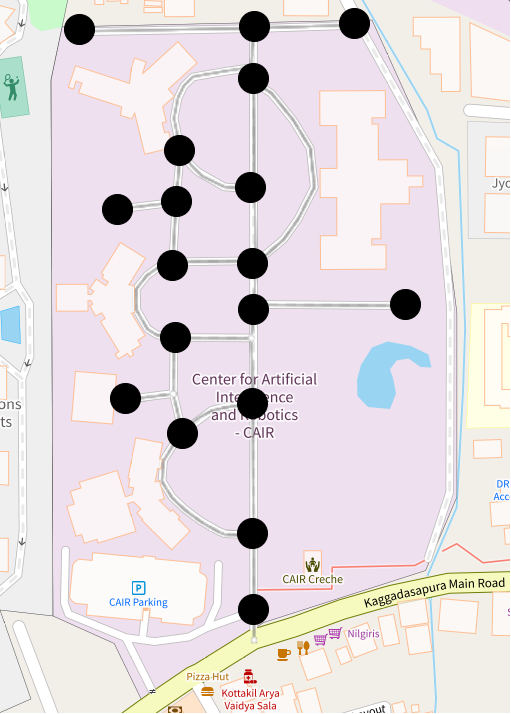}
    \caption{Campus 2}
\end{subfigure}
\begin{subfigure}[b]{0.19\textwidth}
    \includegraphics[width=\textwidth]{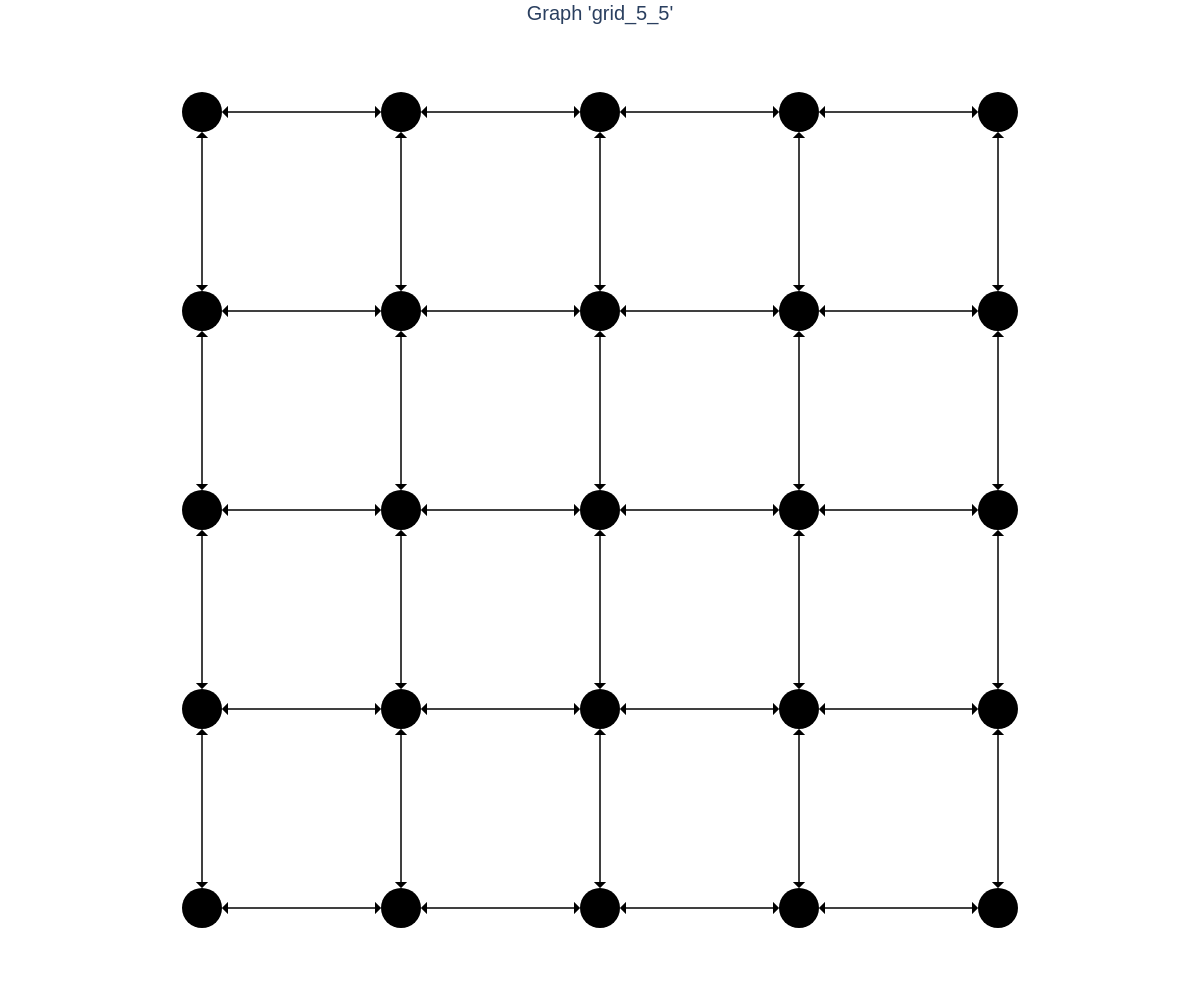}
    \caption{$5 \times 5$ Grid}
\end{subfigure}}
\caption{Different Environments considered for simulation. a) and b) environments mimic the real-world maps.}
\label{fig:Maps}
\end{figure}

Each algorithm setting is evaluated on three different Environments as shown in Figure~\ref{fig:Maps}. All proposed variants of the algorithm are evaluated for various settings (refer to Table \ref{tab:Parameters}). There are 81 unique simulation settings for each variant, these settings are repeated thrice with different initial conditions. Hence, 243 simulations are performed for each variant; therefore, a total of 972 simulations are performed.

\begin{table}[htbp]
    \centering
    \begin{tabular}{|c|c|c|}
    \hline
    Parameter  & Values & Total \\
    \hline
    $|\mathcal{S}|$ & $\{4,5,6\}$ & 3 \\
    $|\mathcal{A}|$ & $\{2,3,4\}$ & 3 \\
    $\mathcal{G}$ & \{IITB, Campus 2, Grid\} & 3 \\
    $H$  & $\{0,3,5\}$ & 3 \\
    \hline
     & Total Settings & 81 \\
    \hline
    \end{tabular}
    \caption{Various simulation settings; $H=0$ signifies two hops instead of three}
    \label{tab:Parameters}
\end{table}

\subsection{Experimental Set-up}
To evaluate the real-time capabilities of PPA variants, we implement them on mobile robots. The experiments are conducted on a 5x5 grid layout. TurtleBot3 Burger robot uses a Raspberry Pi 4B for onboard computing. 
\begin{figure}[htbp]
    \centering
    \fbox{\includegraphics[width = 0.48\textwidth]{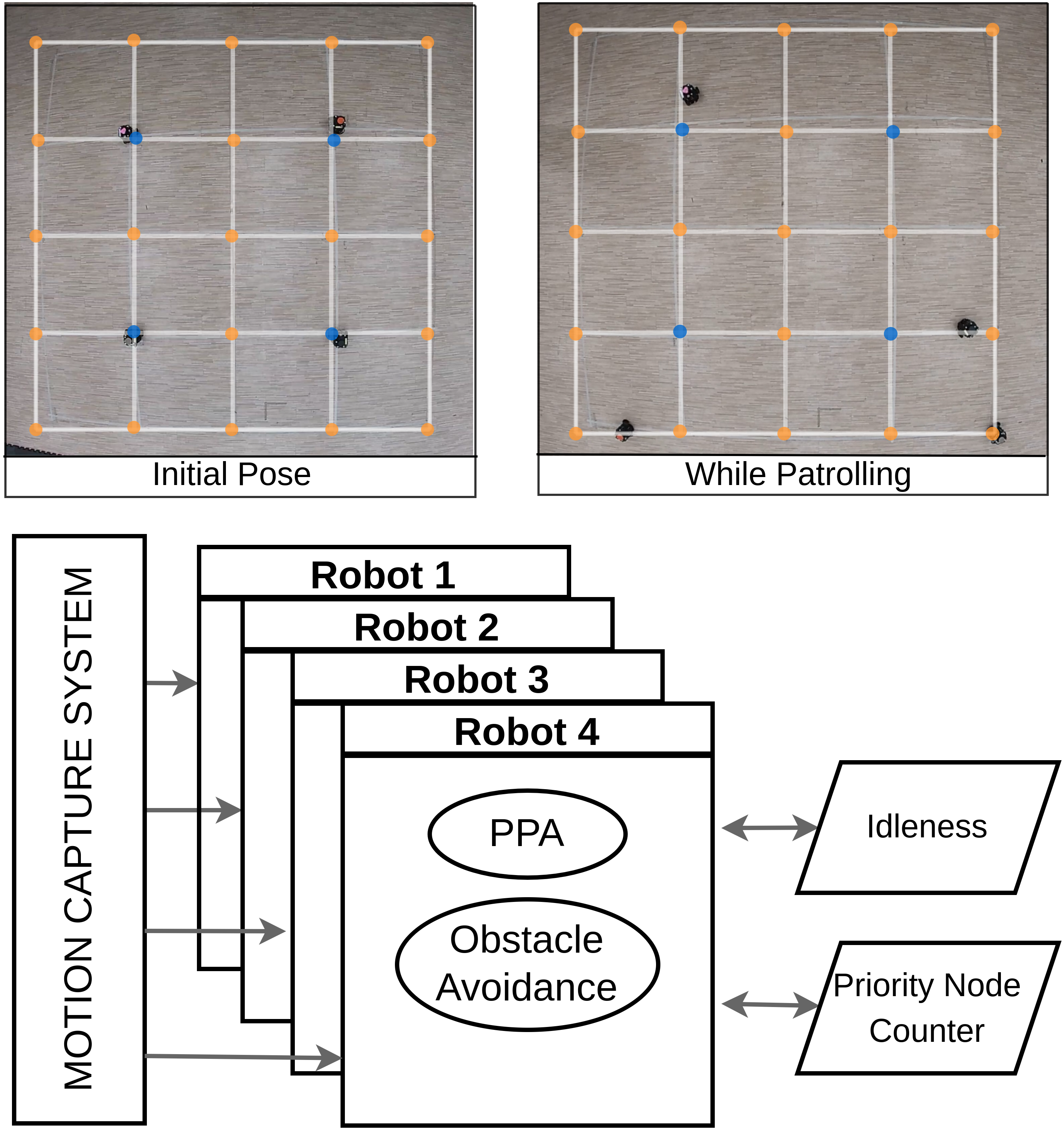}}
    \caption{Experimental setting}
    \label{fig: exp}
\end{figure}
The Raspberry Pi hosts a Ubuntu 20.04 server with ROS Noetic on a quad processor CPU @1.4GHz. Figure \ref{fig: exp} shows the block diagram for experiments.
We implement a hybrid reciprocal velocity obstacle method for multi-robot collision avoidance while patrolling \cite{snape_hybrid_2011}. The robots share the idleness of graphs through ROS topics. Each robot obtains its position on the grid through the motion capture system at IITB-ARMS lab using a ROS package\cite{achtelik_vicon_2023}. Every time a robot arrives at a node, it broadcasts its arrival, and all the robots update their idleness values for the node in the graph. For PPA-Greedy, where a counter is maintained for a visit to each priority node. Robots share information in a similar way through ROS topics. The experimental results explain the memory and computational resources used by the proposed algorithm.  


\subsection{Results and Analysis}
\label{results}

 The performance analysis is presented in a systematic manner using the evaluation metrics namely Priority Nodes' Maximum Idleness, Graph Maximum Idleness, and Idleness Ratio. These evaluation metrics facilitate the validation of our claims on balanced visits in priority patrolling. 

\begin{figure}[htbp]
    \center
    \fbox{\begin{subfigure}[b]{0.48\textwidth}
        \includegraphics[width=\textwidth]{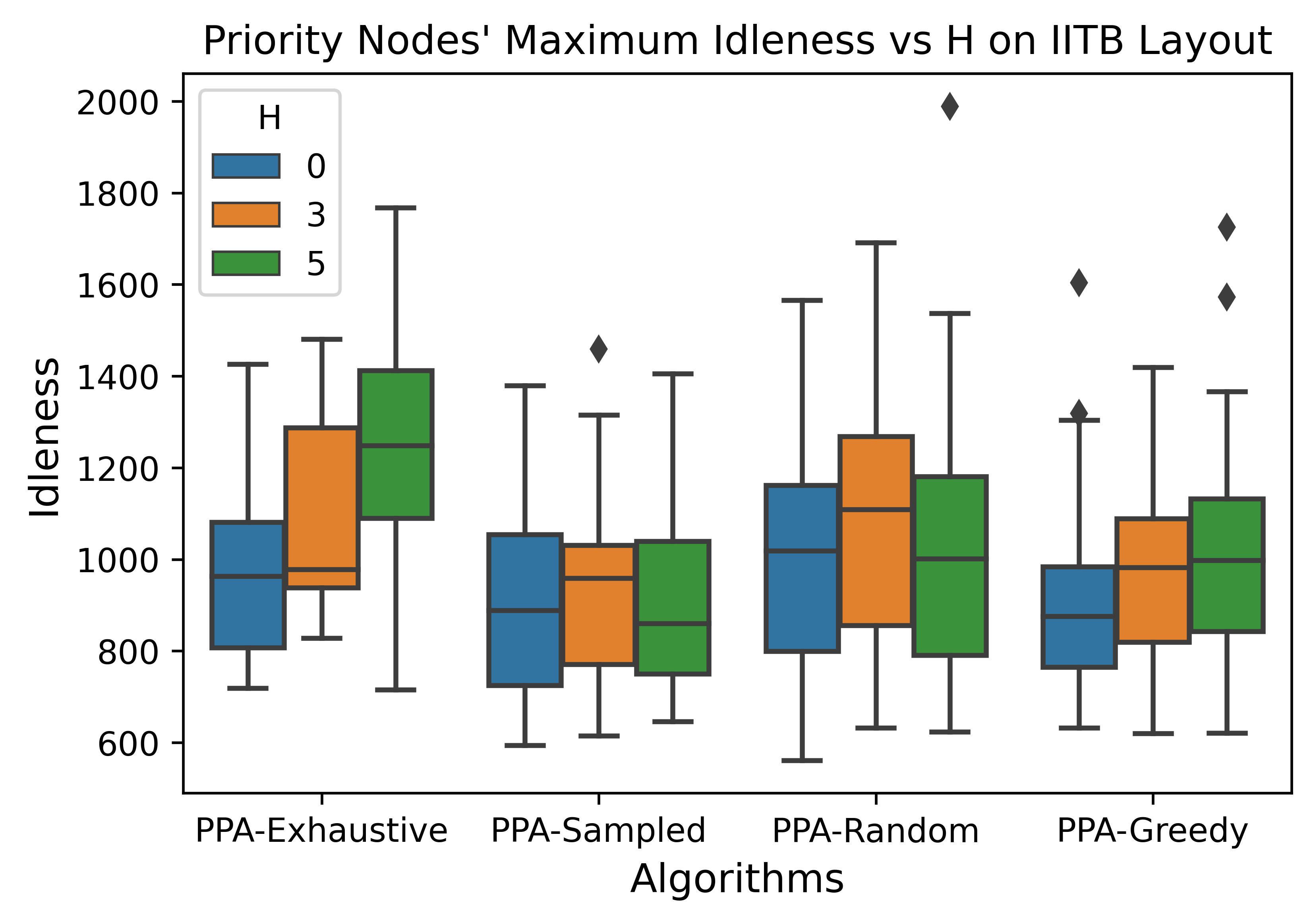}
    \end{subfigure}}
    \fbox{\begin{subfigure}[b]{0.48\textwidth}
        \includegraphics[width=\textwidth]{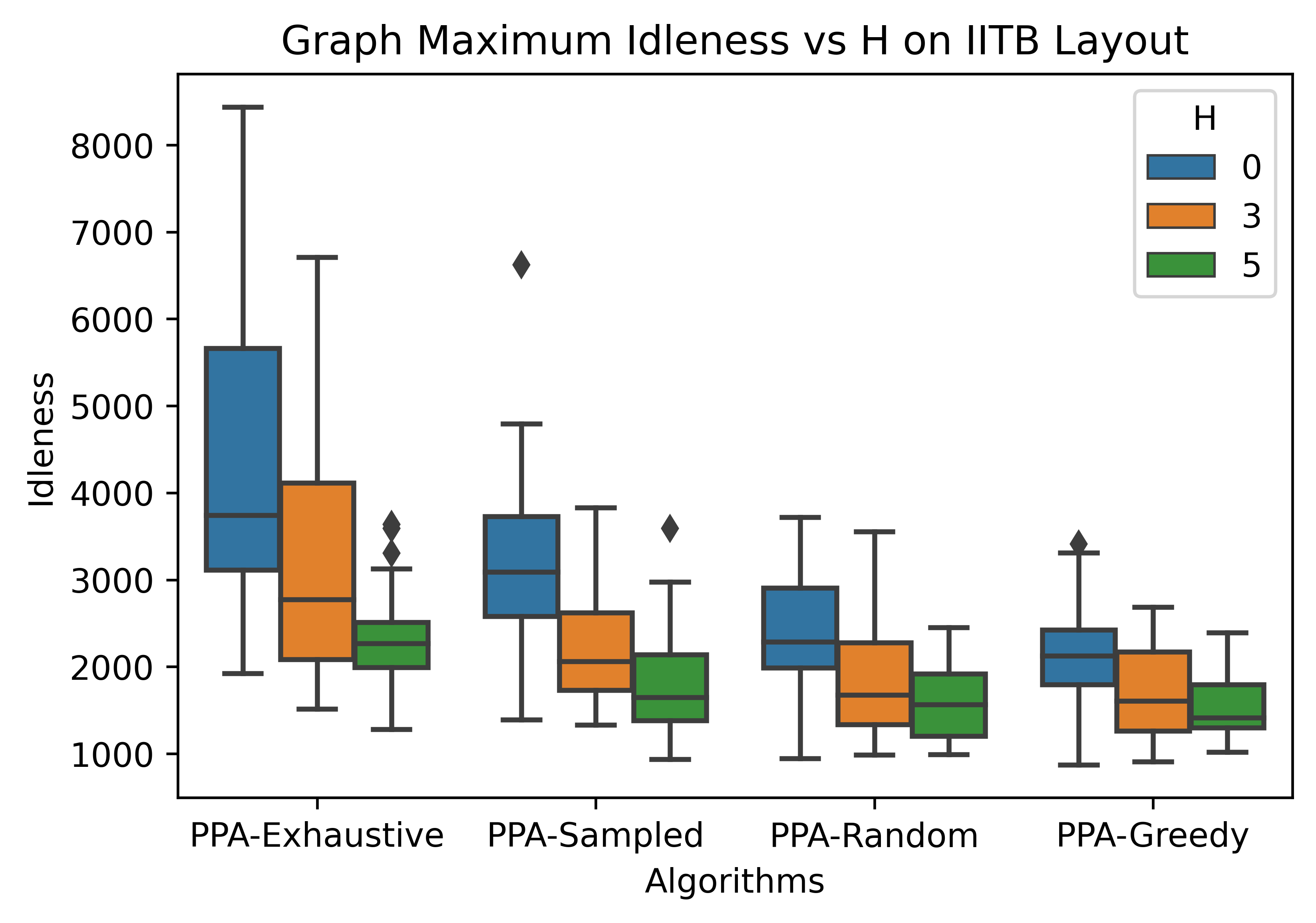}
    \end{subfigure}}
    \caption{Box Plot of Maximum Idleness Values across different PPA variants and Hop values.}
    \label{fig:Hops}
\end{figure}

Figure \ref{fig:Hops} illustrates the maximum Idleness values observed both within Priority Nodes and across the entire node set for simulations conducted on the IIT Bombay (IITB) layout. Notably, under the PPA-Exhaustive variant, we observe a distinct increase in the maximum Idleness values within Priority Nodes as the value of $H$ increases. This trend is attributed to the elongation of Rabbit Walks with a greater magnitude of $H$. Conversely, in the case of other variants, we observe a more even distribution of maximum Idleness values, indicating their robustness across varying $H$, including the case with only two hops ($H=0$).

Across all PPA variants, there is a clear and consistent relationship between the value of $H$ and Graph Maximum Idleness. As the $H$ increases, the Graph's Maximum Idleness value decreases. Notably, the PPA-Greedy variant consistently outperforms other variants in terms of minimizing Graph Maximum Idleness. Under Grid and Campus 2 layouts, PPA-Greedy demonstrates comparable performance with PPA-Exhaustive.

\begin{figure}
    \centering
    \fbox{\begin{subfigure}[b]{0.48\textwidth}
        \includegraphics[width=\textwidth]{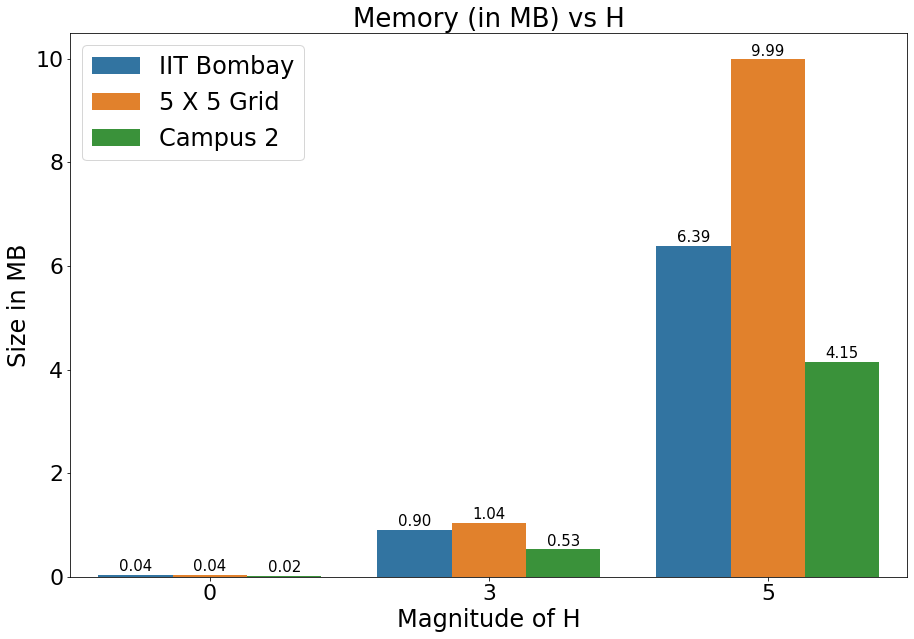}
    \end{subfigure}}
    \caption{Variation of memory used for different graphs and $H$}
    \label{fig: memory}
\end{figure}

Next, we present the results from the experiments.
Figure \ref{fig: memory} illustrates the relationship between memory requirements and $H$ during the generation of Rabbit Walks. The Rabbit Walks generated for all possible pairs of Priority Nodes are considered during the calculation of memory requirements. The number of Priority Nodes is set to four for all layouts.
It is evident from Figure \ref{fig: memory} that the number of Rabbit Walks increases exponentially with $H$. On a similar note, table \ref{tab: Compute} shows the compute time for PPA-Greedy and PPA-Exhaustive with TurtleBot3's onboard computer. The PPA-Exhaustive requires more computational time than PPA-Greedy due to a larger set of walks to parse for assignment.

\begin{table}[htbp]
    \centering
    \begin{tabular}{|c|c|}
    \hline
    Variant & Computational Time \\
    \hline
    PPA\_Greedy & 0.16 \\
    \hline
    PPA\_Exhaustive & 0.64\\
    \hline
    \end{tabular}
    \caption{Various simulation parameters}
    \label{tab: Compute}
\end{table}

\begin{figure}[htbp]
    \center
    \fbox{\begin{subfigure}[b]{0.48\textwidth}
        \includegraphics[width=\textwidth]{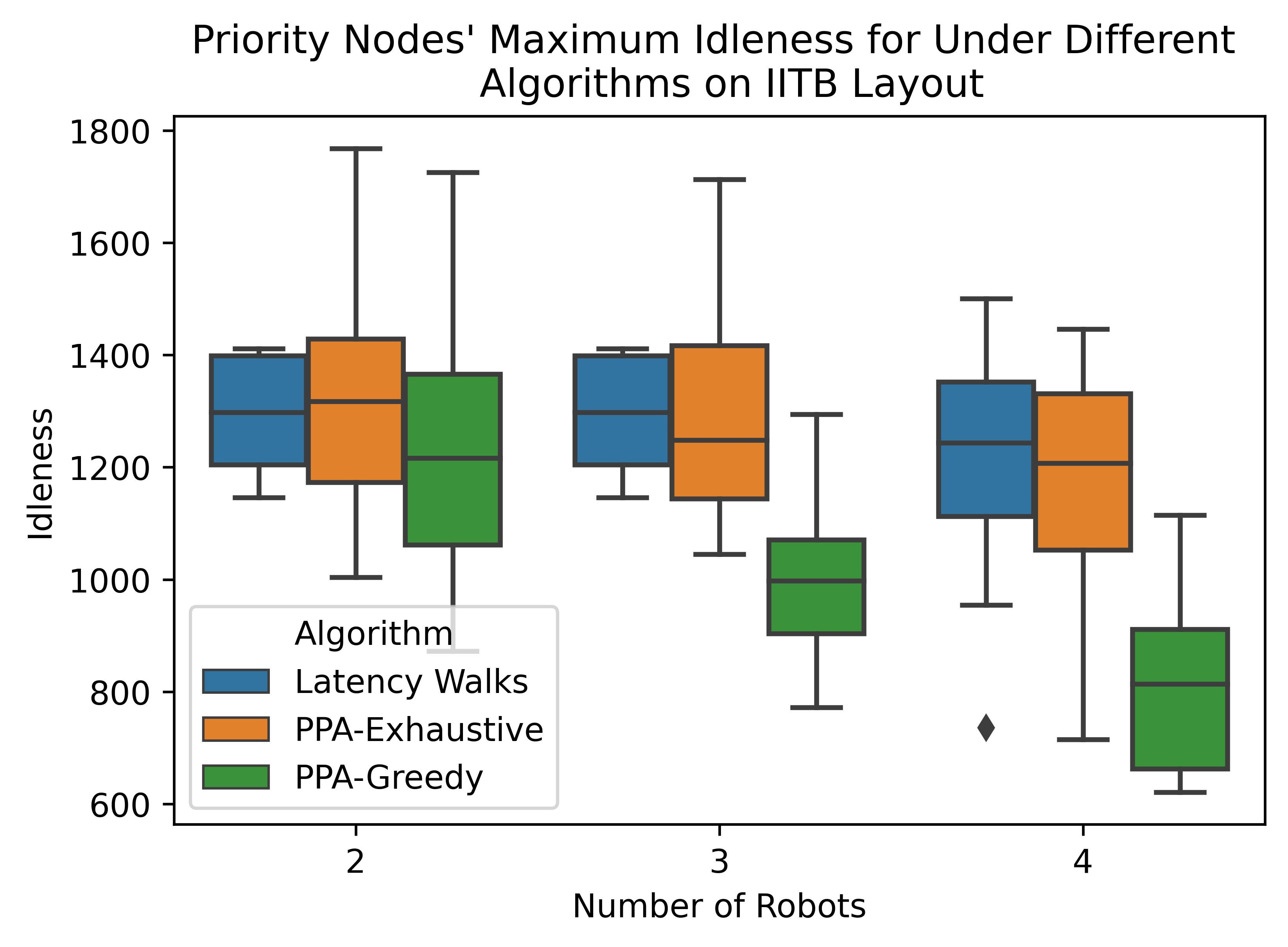}
    \end{subfigure}}
    \fbox{\begin{subfigure}[b]{0.48\textwidth}
        \includegraphics[width=\textwidth]{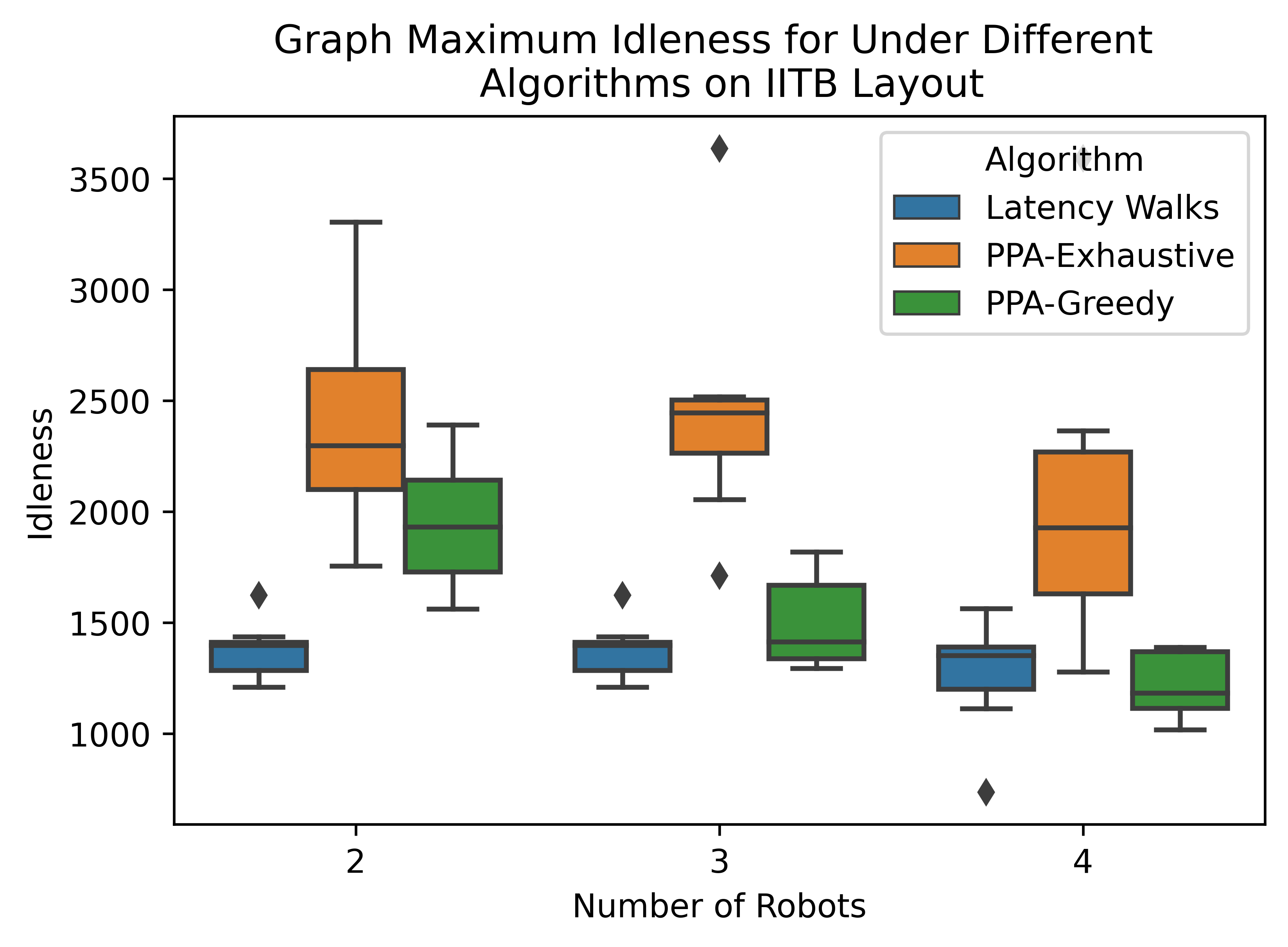}
    \end{subfigure}}
    \caption{Comparison of Maximum Idleness Values across different algorithms and number of robots. ($H = 5$)}
    \label{fig: comp}
\end{figure}

We compare the online performance of our proposed variants (deterministic) with that of an offline method, the Latency Walks Algorithm~\cite{alamdari_min-max_2013}. 
Latency Walks Algorithm uses weighted graphs for priority patrolling. To keep the reference common for comparison, our Latency Walks implementation pre-assigns the weights of the Priority Nodes to the worst Idleness Ratio, while the weights of Non-Priority Nodes are set to 1.
Figure \ref{fig: comp} provides a comparison between PPA-Exhaustive and PPA-Greedy (deterministic variants) when benchmarked against the Latency Walks algorithm, with the weight ratio set to 8, the maximal Idleness Ratio observed across all PPA variants. 

\begin{figure}
    \centering
    \fbox{\begin{subfigure}[b]{0.48\textwidth}
        \includegraphics[width=\textwidth]{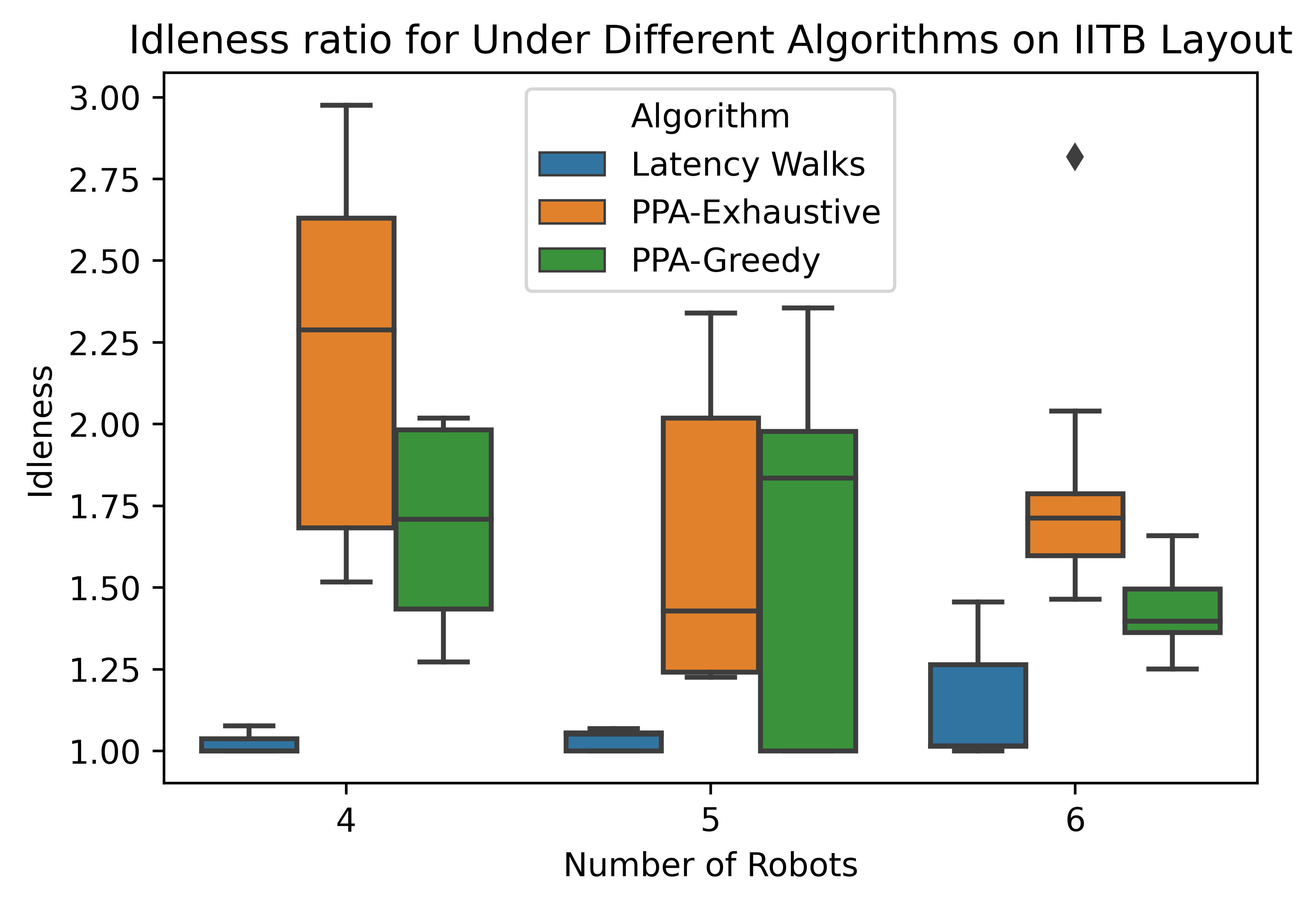}
    \end{subfigure}}
    \caption{Box Plot of Idleness ratio across different PPA variants with $H=5$ }
    \label{fig:idle_ratio}
\end{figure}

Moreover, we evaluate the balance using the Idleness Ratio, a metric that signifies the finite time visits to the Non-Priority nodes during Priority Patrolling. The case of an Idleness Ratio equal to 1 indicates equal priority being given to all nodes.
Figure \ref{fig:idle_ratio} provides a comparison of the idleness ratio among Latency walks and PPA variants with $H = 5$. 

The Latency Walks algorithm consistently exhibits lower values of Graph Maximum Idleness as well as Priority Nodes' Maximum Idleness when compared to both PPA variants for lower $H$ values. Therefore, in memory-scarce systems ($H = 0$ implementation), an offline Priority Patrolling strategy is a solution.
 
With some onboard memory on robots (for example, scenarios with $H=5$) and the PPA-Greedy variant, Graph Idleness values are comparable with an increasing number of agents. Focusing on Priority Nodes, PPA-Greedy outperforms Latency Walks as the number of agents increases. These findings remain consistent across various Graph Layouts.
Whereas the PPA-Exhaustive variant has the opposite effect. It has comparable Priority Nodes' Maximum Idleness values while there is an increase in Graph Maximum Idleness values.
The PPA-Exhaustive variant has a higher range of Idleness Ratios in comparison with PPA-Greedy, thereby creating a greater proportional balance between Priority and Non-Priority.

In summary, results showcase the effectiveness of the PPA-Greedy variant in balancing Graph Maximum Idleness and Priority Maximum Idleness, particularly when compared to PPA-Exhaustive and Latency Walks under various layout scenarios and with varying numbers of agents. These underscore its potential for practical application in real-world patrolling scenarios.


\section{Conclusion and Future works}
\label{conclusion}

While the Patrolling Problem deals with the entirety of the underlying graph at par, the primary objective of the proposed Priority Patrolling Problem is to achieve visits to the Priority Nodes as frequently as possible without compromising the overall patrolling of the environment. In this paper, we develop a novel online practical solution to the Priority Patrolling problem. We show through this work that the suggested algorithm addresses this objective of achieving a proportional balance between patrol of Priority as well as Non-Priority Nodes. The proposed variants are applicable in patrolling scenarios that range from routine inspection which can be handled in a deterministic manner to surveillance and threat detection, which involve arbitrary assignment of the agents' routes during operation. 
The proposed algorithm is independently scalable to (i) onboard resources,(ii) the number of agents, and (iii) graph size. 

Extensions can be explored to investigate different methodologies in terms of the number of Hops while generating Rabbit Walks and complex Reward functions to achieve auxiliary objectives. 
Since the optimal results are a function of the underlying Graph and Agent availability, we are exploring the analytical results on the guarantees of achieving optimality using the observations from this work.

\bibliographystyle{unsrt}
\bibliography{references}

\begin{thebibliography}{10}

\bibitem{pasqualetti_optimal_2010}
Fabio Pasqualetti, Antonio Franchi, and Francesco Bullo.
\newblock On optimal cooperative patrolling.
\newblock In {\em 49th {IEEE} {Conference} on {Decision} and {Control}
  ({CDC})}, pages 7153--7158, Atlanta, GA, USA, December 2010. IEEE.

\bibitem{morales-ponce_optimal_2022}
Oscar Morales-Ponce.
\newblock Optimal patrolling of high priority segments while visiting the unit
  interval with a set of mobile robots.
\newblock {\em Theoretical Computer Science}, 911:1--18, April 2022.

\bibitem{portugal_msp_2010}
David Portugal and Rui Rocha.
\newblock {MSP} algorithm: multi-robot patrolling based on territory allocation
  using balanced graph partitioning.
\newblock In {\em Proceedings of the 2010 {ACM} {Symposium} on {Applied}
  {Computing}}, {SAC} '10, pages 1271--1276, New York, NY, USA, March 2010.
  Association for Computing Machinery.

\bibitem{caccavale_multi-robot_2023}
Riccardo Caccavale, Mirko Ermini, Eugenio Fedeli, Alberto Finzi, Vincenzo
  Lippiello, and Fabrizio Tavano.
\newblock A multi-robot deep {Q}-learning framework for priority-based
  sanitization of railway stations.
\newblock {\em Applied Intelligence}, 53(17):20595--20613, September 2023.

\bibitem{chen_multi-agent_2021}
Jingxi Chen, Amrish Baskaran, Zhongshun Zhang, and Pratap Tokekar.
\newblock Multi-{Agent} {Reinforcement} {Learning} for {Visibility}-based
  {Persistent} {Monitoring}.
\newblock In {\em 2021 {IEEE}/{RSJ} {International} {Conference} on
  {Intelligent} {Robots} and {Systems} ({IROS})}, pages 2563--2570, September
  2021.
\newblock ISSN: 2153-0866.

\bibitem{talmor_power_2017}
Noga Talmor and Noa Agmon.
\newblock On the {Power} and {Limitations} of {Deception} in {Multi}-{Robot}
  {Adversarial} {Patrolling}.
\newblock In {\em Proceedings of the {Twenty}-{Sixth} {International} {Joint}
  {Conference} on {Artificial} {Intelligence}}, pages 430--436, Melbourne,
  Australia, August 2017. International Joint Conferences on Artificial
  Intelligence Organization.

\bibitem{hwang_cooperative_2009}
Kao-Shing Hwang, Jin-Ling Lin, and Hui-Ling Huang.
\newblock Cooperative patrol planning of multi-robot systems by a competitive
  auction system.
\newblock In {\em 2009 {ICCAS}-{SICE}}, pages 4359--4363, August 2009.

\bibitem{malpani_impact_2020}
Rohit Malpani and Madhav Chablani.
\newblock Impact of {CCTV} surveillance on {Crime}.
\newblock {\em Center of Criminal Justice Research}, 2020.

\bibitem{ratcliffe_philadelphia_2011}
Jerry~H. Ratcliffe, Travis Taniguchi, Elizabeth~R. Groff, and Jennifer~D. Wood.
\newblock The {Philadelphia} foot {Patrol} {Experiment}: {A} {Ramdomized}
  controlled {Trial} of {Police} {Patrol} {Effiectiveness} in {Violent} {Crime}
  {Hotspots}.
\newblock {\em Criminology}, 49(3):795--831, August 2011.

\bibitem{halder_robots_2023}
Srijeet Halder and Kereshmeh Afsari.
\newblock Robots in {Inspection} and {Monitoring} of {Buildings} and
  {Infrastructure}: {A} {Systematic} {Review}.
\newblock {\em Applied Sciences}, 13(4):2304, January 2023.

\bibitem{ltd_inspection_2023}
Research {and}~Markets ltd.
\newblock Inspection {Robots} - {Global} {Strategic} {Business} {Report}, 2023.

\bibitem{almeida_recent_2004}
Alessandro Almeida, Geber Ramalho, Hugo Santana, Patrícia Tedesco, Talita
  Menezes, Vincent Corruble, and Yann Chevaleyre.
\newblock Recent {Advances} on {Multi}-agent {Patrolling}.
\newblock In {\em Advances in {Artificial} {Intelligence} – {SBIA} 2004},
  Lecture {Notes} in {Computer} {Science}, pages 474--483, Berlin, Heidelberg,
  2004. Springer.

\bibitem{portugal_survey_2011}
David Portugal and Rui Rocha.
\newblock A {Survey} on {Multi}-robot {Patrolling} {Algorithms}.
\newblock In Luis~M. Camarinha-Matos, editor, {\em Technological {Innovation}
  for {Sustainability}}, {IFIP} {Advances} in {Information} and {Communication}
  {Technology}, pages 139--146, Berlin, Heidelberg, 2011. Springer.

\bibitem{huang_survey_2019}
Li~Huang, MengChu Zhou, Kuangrong Hao, and Edwin Hou.
\newblock A survey of multi-robot regular and adversarial patrolling.
\newblock {\em IEEE/CAA Journal of Automatica Sinica}, 6(4):894--903, July
  2019.

\bibitem{basilico_recent_2022}
Nicola Basilico.
\newblock Recent {Trends} in {Robotic} {Patrolling}.
\newblock {\em Current Robotics Reports}, 3(2):65--76, June 2022.

\bibitem{pippin_performance_2013}
Charles Pippin, Henrik Christensen, and Lora Weiss.
\newblock Performance based task assignment in multi-robot patrolling.
\newblock In {\em Proceedings of the 28th {Annual} {ACM} {Symposium} on
  {Applied} {Computing}}, {SAC} '13, pages 70--76, New York, NY, USA, March
  2013. Association for Computing Machinery.

\bibitem{farinelli_distributed_2017}
Alessandro Farinelli, Luca Iocchi, and Daniele Nardi.
\newblock Distributed on-line dynamic task assignment for multi-robot
  patrolling.
\newblock {\em Autonomous Robots}, 41(6):1321--1345, August 2017.

\bibitem{chen_fast_2012}
Xu~Chen.
\newblock Fast {Patrol} {Route} {Planning} in {Dynamic} {Environments}.
\newblock {\em IEEE Transactions on Systems, Man, and Cybernetics - Part A:
  Systems and Humans}, 42(4):894--904, July 2012.

\bibitem{sea_frequency-based_2018}
Vourchteang Sea, Ayumi Sugiyama, and Toshiharu Sugawara.
\newblock Frequency-{Based} {Multi}-agent {Patrolling} {Model} and {Its} {Area}
  {Partitioning} {Solution} {Method} for {Balanced} {Workload}.
\newblock In {\em Integration of {Constraint} {Programming}, {Artificial}
  {Intelligence}, and {Operations} {Research}}, volume 10848, pages 530--545,
  Cham, 2018. Springer International Publishing.

\bibitem{machado_multi-agent_2003}
Aydano Machado, Geber Ramalho, Jean-Daniel Zucker, and Alexis Drogoul.
\newblock Multi-agent {Patrolling}: {An} {Empirical} {Analysis} of
  {Alternative} {Architectures}.
\newblock In {\em Multi-{Agent}-{Based} {Simulation} {II}}, Lecture {Notes} in
  {Computer} {Science}, pages 155--170, Berlin, Heidelberg, 2003. Springer.

\bibitem{afshani_approximation_2021}
Peyman Afshani, Mark De~Berg, Kevin Buchin, Jie Gao, Maarten Löffler, Amir
  Nayyeri, Benjamin Raichel, Rik Sarkar, Haotian Wang, and Hao-Tsung Yang.
\newblock Approximation {Algorithms} for {Multi}-{Robot} {Patrol}-{Scheduling}
  with {Min}-{Max} {Latency}.
\newblock In {\em Springer {Proceedings} in {Advanced} {Robotics}}, volume~17,
  pages 107--123. Springer International Publishing, 2021.

\bibitem{lauri_two-step_2008}
Fabrice Lauri and Abderrafiaa Koukam.
\newblock A two-step evolutionary and {ACO} approach for solving the
  multi-agent patrolling problem.
\newblock In {\em 2008 {IEEE} {Congress} on {Evolutionary} {Computation}
  ({IEEE} {World} {Congress} on {Computational} {Intelligence})}, pages
  861--868, June 2008.
\newblock ISSN: 1941-0026.

\bibitem{almeida_combining_2003}
Alessandro De~Luna Almeida.
\newblock Combining {Idleness} and {Distance} to {Design} {Heuristic} {Agents}
  for the {Patrolling} {Task}.
\newblock pages 33--40, 2003.

\bibitem{santana_multi-agent_2004}
H.~Santana, G.~Ramalho, V.~Corruble, and B.~Ratitch.
\newblock Multi-agent patrolling with reinforcement learning.
\newblock In {\em Proceedings of the {Third} {International} {Joint}
  {Conference} on {Autonomous} {Agents} and {Multiagent} {Systems}, 2004.
  {AAMAS} 2004.}, pages 1122--1129, July 2004.

\bibitem{velickovic_graph_2018}
Petar Veličković, Guillem Cucurull, Arantxa Casanova, Adriana Romero, Pietro
  Liò, and Yoshua Bengio.
\newblock Graph {Attention} {Networks}.
\newblock International Conference on Learning Representations, February 2018.
\newblock arXiv:1710.10903 [cs, stat].

\bibitem{portugal_cooperative_2016}
David Portugal and Rui~P. Rocha.
\newblock Cooperative multi-robot patrol with {Bayesian} learning.
\newblock {\em Autonomous Robots}, 40(5):929--953, June 2016.

\bibitem{portugal_applying_2013}
David Portugal, Micael~S. Couceiro, and Rui~P. Rocha.
\newblock Applying {Bayesian} learning to multi-robot patrol.
\newblock In {\em 2013 {IEEE} {International} {Symposium} on {Safety},
  {Security}, and {Rescue} {Robotics} ({SSRR})}, pages 1--6, October 2013.
\newblock ISSN: 2374-3247.

\bibitem{alamdari_persistent_2014}
Soroush Alamdari, Elaheh Fata, and Stephen~L Smith.
\newblock Persistent monitoring in discrete environments: {Minimizing} the
  maximum weighted latency between observations.
\newblock {\em The International Journal of Robotics Research}, 33(1):138--154,
  January 2014.

\bibitem{mallya_priority_2021}
Deepak Mallya, Sumanth Kandala, Leena Vachhani, and Arpita Sinha.
\newblock Priority {Patrolling} using {Multiple} {Agents}.
\newblock In {\em 2021 {IEEE} {International} {Conference} on {Robotics} and
  {Automation} ({ICRA})}, pages 8692--8698, Xi'an, China, May 2021. IEEE.

\bibitem{lopez_microscopic_2018}
Pablo~Alvarez Lopez, Evamarie Wiessner, Michael Behrisch, Laura Bieker-Walz,
  Jakob Erdmann, Yun-Pang Flotterod, Robert Hilbrich, Leonhard Lucken, Johannes
  Rummel, and Peter Wagner.
\newblock Microscopic {Traffic} {Simulation} using {SUMO}.
\newblock In {\em 2018 21st {International} {Conference} on {Intelligent}
  {Transportation} {Systems} ({ITSC})}, pages 2575--2582, Maui, HI, November
  2018. IEEE.

\bibitem{snape_hybrid_2011}
Jamie Snape, Jur van~den Berg, Stephen~J. Guy, and Dinesh Manocha.
\newblock The {Hybrid} {Reciprocal} {Velocity} {Obstacle}.
\newblock {\em IEEE Transactions on Robotics}, 27(4):696--706, August 2011.
\newblock Conference Name: IEEE Transactions on Robotics.

\bibitem{achtelik_vicon_2023}
Markus Achtelik.
\newblock Vicon {Bridge}, September 2023.
\newblock original-date: 2013-08-06T09:17:01Z.

\bibitem{alamdari_min-max_2013}
Soroush Alamdari, Elaheh Fata, and Stephen~L. Smith.
\newblock Min-{Max} {Latency} {Walks}: {Approximation} {Algorithms} for
  {Monitoring} {Vertex}-{Weighted} {Graphs}.
\newblock In {\em Algorithmic {Foundations} of {Robotics} {X}}, Springer
  {Tracts} in {Advanced} {Robotics}, pages 139--155, Berlin, Heidelberg, 2013.
  Springer.

\end{thebibliography}
\end{document}